\documentclass[10pt,twocolumn,letterpaper]{article}

\usepackage{iccv}
\usepackage{times}
\usepackage{epsfig}
\usepackage{graphicx}
\usepackage{amsmath}
\usepackage{amssymb}

\usepackage[pagebackref=true,breaklinks=true,letterpaper=true,colorlinks,bookmarks=false]{hyperref}

 \iccvfinalcopy % *** Uncomment this line for the final submission

\ificcvfinal\fi
\long\def\comment#1{}

\begin{document}
	
	%TODO: weakly means with out fully labeled! ?
	
	\title{Semi and Weakly Supervised Semantic Segmentation Using Generative Adversarial Network}
	
	\author{Nasim Souly\\
		Center for Research in Computer Vision(CRCV)\\
         University of Central Florida \\
		{\tt\small nsouly@eecs.ucf.edu}\\
		\and
		Concetto Spampinato \\
		University of Catania\\
		{\tt\small cspampin@dieei.unict.it}
		\and 
		Mubarak Shah\\
     	Center for Research in Computer Vision(CRCV)\\
         University of Central Florida\\
		{\tt\small shah@crcv.ucf.edu}
	}

	\maketitle
	% \MS{I thought previous title had also Weakly supervised, did  you want to remove weakly?}
	
\begin{abstract}
		Semantic segmentation has been a long standing challenging task in computer vision. It aims at assigning a label to each image pixel and needs significant number of pixel-level annotated data, which is often unavailable. To address this lack, in this paper, we leverage, on one hand, massive amount of available unlabeled or weakly labeled data, and on the other hand, non-real images created through Generative Adversarial Networks. In particular, we propose a semi-supervised framework -- based on Generative Adversarial Networks (GANs) -- which consists of a generator network to provide extra training examples to a multi-class classifier, acting as discriminator in the GAN framework, that assigns sample a label $\textup{y}$ from the $K$ possible classes or marks it as a fake sample (extra class). The underlying idea is that adding large fake visual data forces real samples to be close in the feature space, enabling a bottom-up clustering process, which, in turn, improves multiclass pixel classification. To ensure higher quality of generated images for GANs with consequent improved pixel classification, we extend the above framework by adding weakly annotated data, i.e., we provide class level information to the generator.
    	We tested our approaches on several challenging benchmarking visual datasets, i.e. PASCAL, SiftFLow, Stanford and CamVid, achieving competitive performance also compared to state-of-the-art semantic segmentation methods.
		%TODO: add camvid
\end{abstract}
	
\section{Introduction}
Semantic segmentation, i.e., assigning a label from a set of classes to each pixel of the image, is one of the most challenging tasks in computer vision because of the high variation in appearance, texture, illumination, etc. of visual scenes as well as multiple viewpoints and poses of different objects. Nevertheless, despite the enormous work in past years \cite{chen2014semantic}, \cite{long2015fully}, it is still not fully solved, even though recent deep methods have demonstrated to be a valuable tool. 
However, deep networks require large annotated visual data that, in the case of semantic segmentation, should be at the pixel-level (i.e., each {\em pixel} of training images must be annotated), which is highly prohibitive to obtain.
	
% 	%TODO : intro
	\begin{figure}[t]
		\begin{center}
			\includegraphics[height =0.44\textwidth]{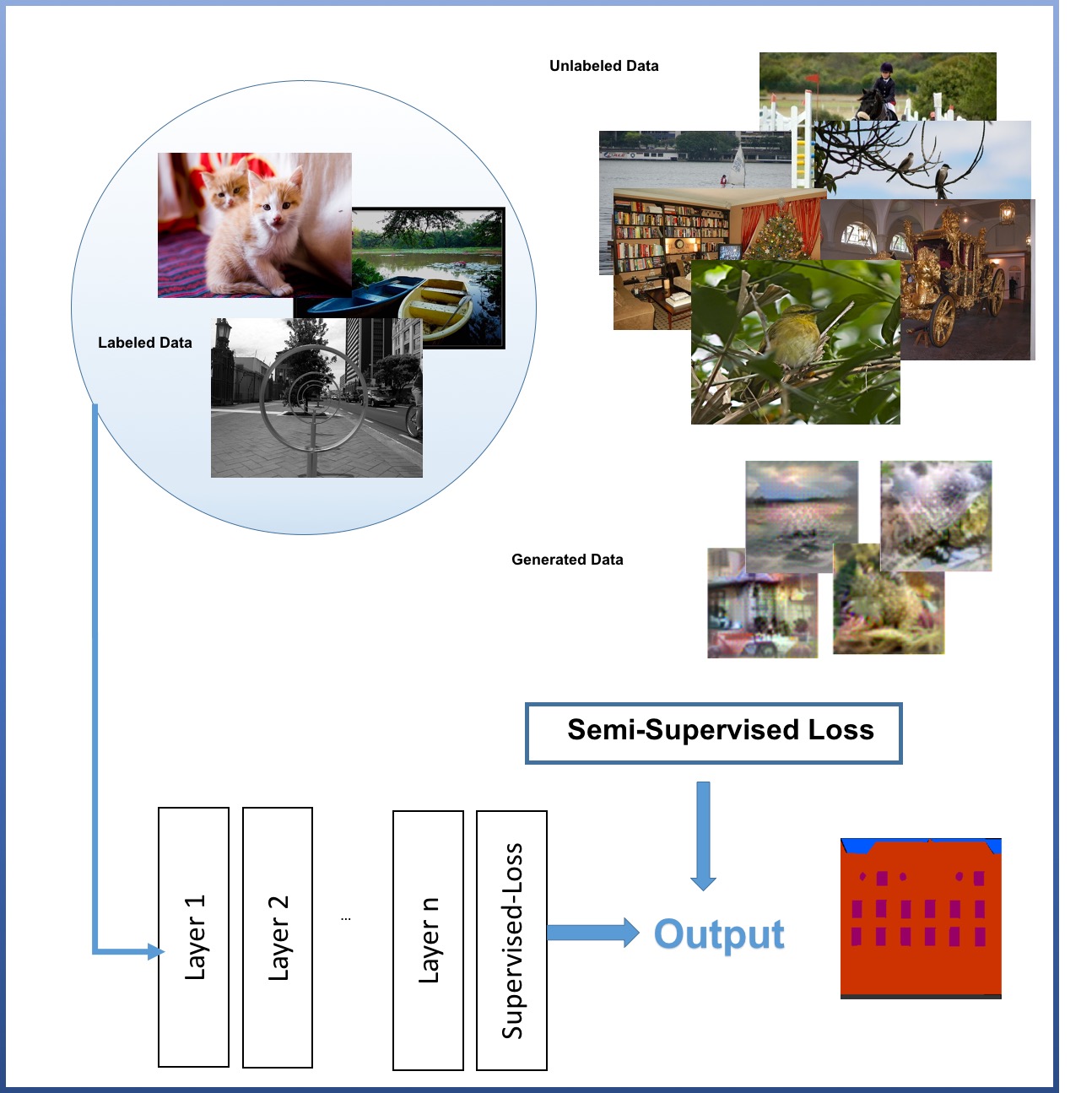}
		\end{center}
		\caption{Our idea is to employ a small set of labeled data together with large available unlabeled data (both realistic and fake) to identify hidden patterns supporting semantic segmentation.}
% 			\vspace{-5mm}
		\label{fig:intro2}
	
	\end{figure}
	
An alternative to supervised learning is unsupervised learning leveraging large amount of available unlabeled visual data. Unfortunately {\em unsupervised} learning methods have not been very successful for semantic segmentation, because they lack the notion of classes trying merely to identify consistent regions and/or region boundaries \cite{valpola2015neural}.
	
{\em Semi-Supervised } Learning (SSL) is halfway between supervised and unsupervised learning, where in addition to unlabeled data, some supervision is also given, e.g., some of the samples are labeled. In semi-supervised learning, the idea is to identify some specific hidden structure -- $p(x)$ from unlabeled data $x$ --under certain assumptions - that can support classification $p(y|x)$, with $y$ class label.
In this paper, we aim to leverage unlabeled data to find a data structure that can support the semantic segmentation phase , as shown in Fig. \ref{fig:intro2}.
In particular, we exploit the assumption that if two data points $x_{1},x_{2}$ are close in the input feature space, then the corresponding outputs (classifications) $y_{1}, y_{2}$ should also be close (smoothness constraint) \cite{chapelle2009semi}.
This concept can be applied to semantic segmentation, i.e., pixels lying on the same manifold should be close in the label, thus should be classified in the same class. This means that unsupervised data acts as regularizer in deep networks, thus improving their generalization capabilities. 
	
Under the above assumption, in this paper, we employ generative adversarial networks (GANs) \cite{goodfellow2014generative} to support semi-supervised segmentation by generating additional information useful for the classification task. 
GANs have, recently, gained a lot of popularity because of their ability in generating high-quality realistic images with several, documented, advantages over other traditional generative models \cite{huszar2015not}. 
In our GAN-based semi-supervised semantic segmentation method, the generator creates large realistic visual data that, in turn, forces the discriminator to learn better features for more accurate pixel classification. 
Furthermore, to speed up and improve the quality of generated samples for better classification, we also condition the GANs with additional information -- weak labels -- on image classes. 
In our formulation of GAN, we employ a generator network similar to \cite{radford2015unsupervised}, which, given a noise vector as an input, generates an image to be semantically segmented by b) a multiclass classifier (our discriminator) that, in addition to classifying the pixels into different semantic categories, determines whether a given image belongs to training data distribution or is coming from a generated data. 
	
The performance analysis over several benchmarking datasets for semantic segmentation, namely Pascal VOC 2012, SiftFlow, StanfordBG, and CamVid, shows the effectiveness of our approach compared to state-of-the-art methods. 
	
Summarizing, the main contributions of this paper are:
\begin{itemize}
	\item We present a GAN network framework which extends the typical GAN to pixel-level prediction and its application in semantic segmentation.
	\item Our network is trained in semi-supervised manner to leverage from generated data and unlabeled data.
	\item Finally, we extend our approach to weakly supervised learning by employing conditional GAN and available image-level labeled data.
\end{itemize}
	
%I USUALLY DON'T PUT THIS IN PAPERS. IT DOES NOT ADD ANYTHING IMPORTANT AND WASTES SOME SPACE WHICH MAY BE NEEDED FOR CONVEYING MORE IMPORTANT MESSAGES. 
	
The organization of the rest of the paper is as follows. In the next section, we review recent methods for semantic segmentation. In Section 3, we present our approach, where we first provide a brief background of generative adversarial networks, then we describe the design and structure of our proposed models for both semi-supervised and weakly-supervised learning. This is followed by System Overview related to training and inference, which is covered in section 4. Section 5 deals with experimental results, where we report our results on Pascal VOC 2012, SiftFlow, StanfordBG and CamVid datasets. Finally, we conclude the paper in Section 6. 
	
\section {Related Work}
Semantic segmentation has been widely investigated in past years. Some of the existing methods aim at finding a graph structure over the image, by using  Markov Random Field (MRF) or Conditional Random Field (CRF),  to capture the context of an image as well as using classifiers to label different entities (pixels, super pixels or patches) \cite{tighe2014scene} \cite{guo2014labeling} \cite{souly2016scene}. Additional information, such as long range connections, to refine further the segmentation results have been also proposed \cite{souly2016scene}. Nonetheless, these methods employ hand crafted features for classification, which makes them hardly generalizable.

Convolutional Neural Networks (CNNs) have been very popular recently in many computer vision applications including semantic segmentation. For instance, \cite{mostajabi2015feedforward} and \cite{farabet2013learning} leverage deep networks to classify super-pixels and label the segments. More recent methods such as \cite{long2015fully} apply per-pixel classification using a fully convolutional network. This is achieved  by transforming fully-connected layers of CNN (VGG16) into convolutional layers and using the pre-trained ImageNet model to initialize the weights of the network. Multiple deconvolution layers \cite{noh2015learning} have been also employed to enhance pixel classification accuracy. 
Post-processing based on MRF or CRF on top of deep network framework has been adopted, as in \cite{chen2014semantic}, to refine pixel label predictions. For example, in \cite{schwing2015fully} the error of MRF inference is passed backward into CNN in order to train jointly CNN and MRF. However, this kind of post-processing is rather expensive since, for each image during training, iterative inference should be performed.
	
The aforementioned methods are based on supervised learning and rely strongly on large annotated data, which is often unavailable. To cope with this limitation, a few number of weakly or semi-supervised semantic segmentation methods have been proposed,\cite{papandreou2015weakly}, \cite{pathak2014fully}, \cite{dai2015boxsup}. These approaches assume that weak annotations (bounding boxes or image level labels) are available during training and that such annotations, combined with limited pixel-level labels, force deep networks to learn better visual features for classification. In \cite{hong2015decoupled}, the authors address the semantic segmentation as two separate tasks of classification and segmentation, and assume image level labels for all images in data set and a limited number of fully pixel-level labeled data are available.
%% WHAT ARE THE LIMITATIONS OF THE ABOVE METHODS?
	
To tackle the limitations of current methods, we propose to use GANs in semi-supervised learning for semantic segmentation to leverage freely available data and additional synthetic data to improve the fully supervised methods. 
While generative methods have been largely employed in unsupervised and semi-supervised learning for visual classification tasks \cite{springenberg2015unsupervised}, \cite{salimans2016improved}, very little has been done for semantic segmentation, e.g., \cite{luc2016semantic}. In particular, \cite{luc2016semantic} aims at creating probability maps for each class for a given image, then the discriminator has to distinguish between generated maps and ground truth. 
Our method is significantly different from it as 1) we handover the discriminator to find the labels of pixels, 2) we leverage unlabeled data along side generated data, in an adversarial way, to compete in getting realistic labels, and 3) we use conditional GAN to enhance the quality of generated samples for better segmentation performance as well as to make GAN training more stable.
	
\begin{figure*}[t]
    \begin{center}
	    \includegraphics[height=0.28\textwidth]{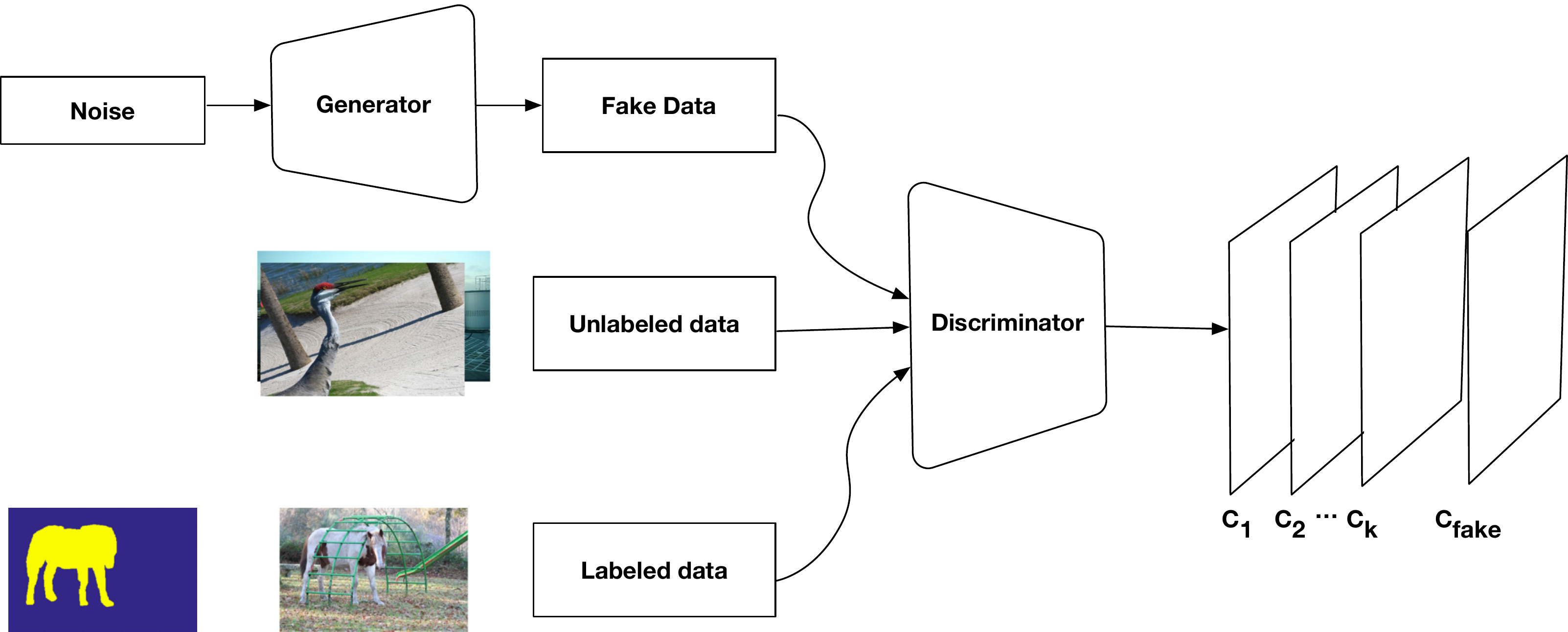}
	\end{center}
	\caption{Our semi-supervised convolutional GAN architecture. Noise is used by the Generator to generate an image. The Discriminator uses generated data, unlabeled data and labeled data to learn class confidences and produces confidence maps for each class as well as a label for a fake data. }
% 		\vspace{-5mm}
\label{fig:semi}
\end{figure*}
% 	\vspace{-3mm}
	
\section{Proposed Approach}
In this section, first we briefly cover the background about GANs and then present our network architectures and corresponding losses for semi and weakly supervised semantic segmentation.
\subsection{Background}
\subsubsection{ Generative Adversarial Network}

Generative Adversarial Network (GAN) is a framework introduced by \cite{goodfellow2014generative} to train deep generative models. It consists of a generator network, $G$, whose goal is to learn a distribution, $p_z$ matching the data, and a discriminator network $D$, which tries to distinguish between real data (from true distribution $p_{data}(x)$) and fake data (generated by the generator). $G$ and $D$ are competitors in a minmax game with the following formulation:
\begin{equation}
    \begin{array}{l}
	\underset{G}{\textup{min}}\ \underset{D}{\textup{max}} \ V(D,G)=\mathbb{E}_{x\sim p_{data}(x)}[log(D(x))]+ \\
	\mathbb{E}_{z\sim p_{z}(z)}[log(1-D(G(z))],
	\end{array} 
	\label{eq:gan}
\end{equation}

where $\mathbb{E}$ is the empirical estimate of expected value of the probability. 
G transforms a noise variable $z$ into $G(z)$, which basically is a sample from distribution $p_{z}$, and ideally distribution $p_{z}$ should converge to distribution $p_{data}$. Minimizing $ \textup{log}(1-D(G(z))$ is equivalent to maximizing $\textup{log}(D(G(z))$, and it has been shown that it would lead to better performance, so we follow the latter formulation. 
	
\subsection{Semi Supervised Learning using Generative Adversarial Networks}

In semi-supervised learning, where class labels (in our case pixel-wise annotations) are not available for all training images, it is convenient to leverage unlabeled data for estimating a proper prior to be used by a classifier for enhancing performance. 
In this paper we adopt and extend GANs, to learn the prior fitting the data, by replacing the traditional discriminator $D$ with a fully convolutional multiclass classifier, which, instead, of predicting whether a sample $\textup{x}$ belongs to the data distribution (it is real or not), it assigns to each input image pixel a label $\textup{y}$ from the $K$ semantic classes or mark it as a fake sample (extra $K+1$ class).  More specifically, our discriminator $D(x)$ is a function parametrized as a network predicting the confidences for $K$ classes of image pixels and softmax is employed to obtain the probability of sample $\textup{x}$ belonging to each class. In order to be consistent with GAN terminology and to simplify notations we will not use $D_k$ and use $D$ to represent pixel-wise multi-class classifier. Generator network, $G$, of our approach maps a random noise $z$ to a sample $G(z)$ trying to make it similar to training data, such that the output of $D$ on that sample corresponds to one of the real categories. $D$ is, instead,  trained in order to label the generated samples $G(z)$ as fake. Figure \ref{fig:semi} provides a schematic description of our semi-supervised convolutional GAN architecture and shows that we feed three inputs to the discriminator: labelled data, unlabelled data and  fake data. Accordingly, we define a pixel-wise discriminator loss, $\mathcal{L}_{D}$, in order to account for the three kind of input data, as follows: 

{ \small \begin{equation}
		\begin{array}{l}
		\mathcal{L}_{D} =\underset{D}{\textup{max}}\ \mathbb{E}_{x\sim p_{data}(x)}[log(D(x))] \\
		-\gamma \mathbb{E}_{x,y\sim p(y,x)}[\textup{CE}(y,P(y|x,D))]
		+\mathbb{E}_{z\sim p_{z}(z)}[log(1-D(G(z))],
		\end{array} 
		\label{eq:semiminmax}
		\end{equation} }
where
\begin{equation} 
D(x) = \textup [1-P(y = fake|x) ]. \\
\end{equation} 
	
with $y=1 \cdots K$ being the semantic class label, $p(x,y)$ the joint probability of labels ($y$) and data ($x$), $CE$ the cross entropy loss between labels and probabilities predicted by $D(x)$. The first term of $\mathcal{L}_{D}$ is devised for unlabeled data and aims at decreasing the probability of pixels belonging to the fake class. 
The second term accounts for all pixels in labeled data to be correctly classified in one of the $K$ available classes, while the third loss term aims at driving the discriminator in distinguishing real samples from fake ones generated by $G$.

The generator loss, $\mathcal{L}_G$ is defined as follows:
	
\begin{equation} 
	\mathcal{L}_G = \underset{G}{\textup{min}} \ \mathbb{E}_{z\sim p_{z}(z)}[log(1-D(G(z))]
\label{Gloss}
\end{equation} 

It can be noted that our GAN formulation is different from
typical GANs, where the discriminator is a binary classifier for discriminating real/fake images, while our discriminator performs multiclass pixel categorization.%, as shown by the  pixel-wise loss in Equation \ref {eq:semiminmax}.

\subsection{ Weakly Supervised Learning using Conditional Generative Adversarial Networks}

An recent extension of GANs is conditional GANs \cite{mirza2014conditional}, where generator and discriminator are provided with extra information (e.g., image class labeles) to driving the generator. The traditional loss function, in this case, becomes:
\begin{equation}
	\begin{array}{l}
	\underset{G}{\textup{min}}\ \underset{D}{\textup{max}} \ V(D,G)=\mathbb{E}_{x,l\sim p_{data}(x,l)}[log(D(x,l))]+ \\
	\mathbb{E}_{z\sim p_{z}(z,l) , l\sim p_{l}(l)}[log(1-D(G(z,l),l)], 
	\end{array} 
	\label{eq:cgan}
	\end{equation}
where $p_{l}(l)$ is the prior distribution over class labels, $D(x,l)$ is joint distribution of data, $x$, and labels $l$ and $G(z,l)$ is joint distributions of generator noise $z$ and labels $l$ indicating that labels $l$ control the conditional distribution of $p_{z}(z|l)$ of the generator.
	
Semantic segmentation can naturally fit in this model, as long as additional information on training data is available, e.g., image level labels (whose annotation is much less expensive than pixel level one). We use this side-information on image classes to train our GAN network with weak supervision.
The rationale of exploiting weak supervision in our framework lies on the assumption that when image classes are provided to the generator, it is forced to learn co-occurrences between labels and images resulting in higher quality generated images, which, in turn, help our multiclassifier to learn more meaningful features for pixel-level classification and true relationships between labels.

\begin{figure*}[t]
		\begin{center}
			\includegraphics[height=.28 \textwidth]{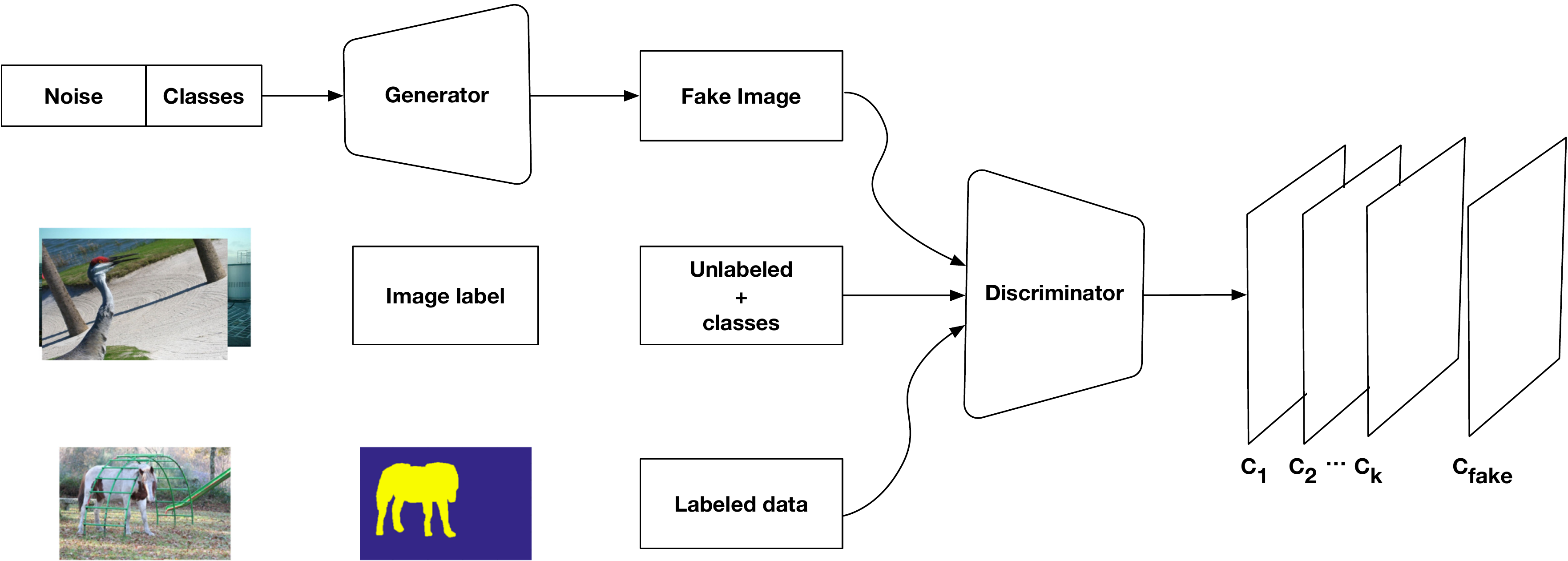}
		\end{center}
		\caption{Our weakly-supervised convolutional GAN architecture. In addition to noise, class label information is used by the Generator to generate an image. The Discriminator uses generated data, unlabeled data plus image-level labels and pixel-level labeled data to learn class confidences and produces confidence maps $C_1,C_2,\ldots, C_k$ for each semantic class as well as a label $C_{fake}$ for the fake data. }
% 			\vspace{-3mm}
		\label{fig:weak}
\end{figure*}

Our proposed GAN network architecture for weakly supervised semantic segmentation is shown in Figure \ref{fig:weak}. The discriminator is fed with unlabeled images together with class level information, generated  images coming from $G$ and pixel-level labeled images. Thus, the discriminator loss, $\mathcal{L}_D$, is comprised by three terms: the term for weakly labeled sample data belonging to data distribution $p_{data}(x,l)$, the term for loss of generated samples not belonging to the true distribution, and the term for the loss of pixels in labeled data classified correctly. Hence, the discriminator loss $\mathcal{L}_D$ is as follows:
	\begin{equation}
	\begin{array}{l}
	% L_{D} =- \mathbb{E}_{x,l\sim p_{data(x,l)}}\textup{ log} [p(y \in K_{i} \subset{1...K}|x) ] \\
	% - \mathbb{E}_{x,l \sim p_{z,l}(x,l)}\textup{ log} [p(y = fake|x) ]\\-\gamma \mathbb{E}_{x,y\sim p(y,x)}[\textup{CE}(y,P(y|x,D))],
	\mathcal{L}_{D} = \underset{D}{\textup{max}} \ \mathbb{E}_{x,l\sim p_{data(x,l)}}\textup{ log} [p(y \in K_{i} \subset{1...K}|x) ] \\
	+  \mathbb{E}_{x,l \sim p_{z,l}(x,l)}\textup{ log} [p(y = fake|x) ] \\ -\gamma \mathbb{E}_{x,y\sim p(y,x)}[\textup{CE}(y,P(y|x,D))],
	
	\end{array} 
	\label{eq:condgan}
	\end{equation}
where $K_{i}$ indicates the classes present in the image. Here we have modified the notations for probability distributions and expectation to include label $l$. Conditioning space $l$ (labeled) in loss $L_D$ aims at  controlling the generated samples, i.e., given image classes along with the noise vector the generator attempts to maximize the probability of seeing labels in the generated images, while the goal of discriminator is to suppress the probability of real classes for generated data and encourage high confidence of image level labels for unlabeled data. The generator loss is similar to the one used for semi-supervised case (see Eq. \ref{Gloss}), and aims at enforcing the image-level labels to be present in the generated images. For unlabeled data, we use negative log-likelihood of confidences, favoring the labels occur in the image, meaning that we add a fixed value to pixel confidences for image-level labels.  
%The whole architecture for our weakly-supervised learning GAN for semantic segmentation is shown in Figure \ref}.
% TODO: defining unlabeled loss (weak label loss) based on pseudo label

% 	\vspace{-2mm}
	%TODO: why only gen
\section{System Overview}
	
\begin{figure}
	\begin{center}
			\includegraphics[width=.45\textwidth]{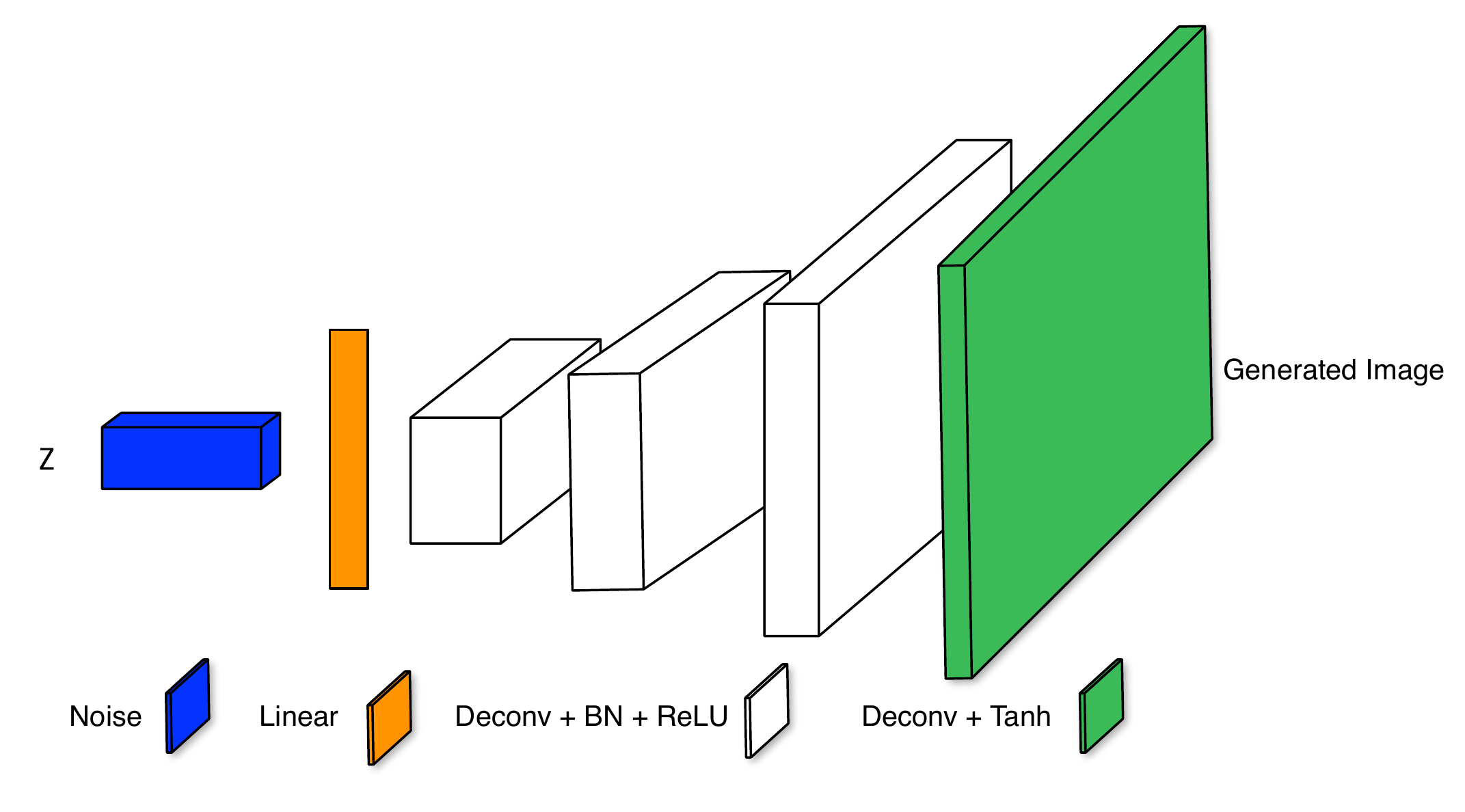}
		\end{center}
		\caption{The generator network of our GAN architecture. The noise is a vector of size 100 sampled from a uniform distribution. The number of feature maps in the five different convolutional layers, respectively, are 769, 384, 256, 192 and 3. }
% 			\vspace{-6mm}
		\label{fig:gen}
\end{figure}

In this section, we present the details of our deep networks, including the discriminator (classifier) and the generator. In both settings, i.e., semi-supervised and weakly-supervised approaches, the discriminator is a fully convolutional network \cite{long2015fully} using VGG16 convolutional layers plus 1 or 3 deconvolution layers, which generates $K+1$ confidence maps. The generator, instead, consists of 4 deconvolution layers transforming noise (and noise plus image class information) into an image (see Fig.~\ref{fig:gen}). 
% As conditioning signals for the generator in our weakly-supervised scenario we employed ``one-hot'' class vectors. 

	The generator network, shown in Figure \ref{fig:gen}, starts with noise, followed by a series of deconvolution filters and generates a synthetic image resembling samples from real data distribution. The generator loss enforces the network to  minimize the distance between $D(G(z_{i}))$ and $y_i \in {l_{i}...l_{K}}$, as shown in Equation \ref{eq:semiminmax}. 
	
% 	In a typical GAN model, the discriminator's objective is to distinguish between the true data distribution and the distribution of the generated (fake) samples. In semi-supervised learning, this definition is extended to determine whether the data is from one of the classes or is generated from a noise.
	The discriminator loss %shown in (figure \ref{fig:dis}),
	is the sum of cross entropy between labeled data and the output of classifiers. This enforces that the discriminator should classify pixels from the generated image (data) into the fake class and unlabeled data to the true classes.
	
	% TODO: remove these 2
	
	% \begin{figure*}[t]
	% \begin{center}
	%  \includegraphics[width=.85\textwidth]{./discrimator_fig.pdf}\\
	%   %\includegraphics[width=0.8\linewidth]{egfigure.eps}
	% \end{center}
	%   \caption{Our discriminator network in GAN architecture. The network is based on a fully convolutional VGG 16 network with deconvolution layers.  }
	% \label{fig:dis}
	% \end{figure*}
	
	% \begin{figure}[t]
	% \begin{center}
	%  \includegraphics[width=.46\textwidth]{./ploterrors.pdf}
	%   %\includegraphics[width=0.8\linewidth]{egfigure.eps}
	% \end{center}
	%   \caption{Behavior of different losses for  VOC data set using $30\%$ (roughly 400 images) of labeled data via weakly supervised approach trained for 100 epochs. Top Left shows the  overall Semantic Segmentation Loss. The remaining three losses correspond to three terms in equation \ref{eq:semiminmax}.}
	% \label{fig:errplot}
	% \end{figure
	In weakly supervised training, we impose the constraint on the generator that, instead, of generating generic images from data distribution, it produces samples belonging to specific visual classes provided as input to it.
% 	TODO: i will put the result for this in supplementary  these are for one-hot vector
    % To do that, we add a linear layer followed by leaky-ReLU to the generator, which compresses a feature vector of the image to a $K$ dimensional vector, and then this vector is concatenated to the noise sampled from the noise distribution.
    To do that, a one-hot image classes vector is concatenated to the noise sampled from the noise distribution.
    Afterward, the deconvolution layers are applied similar to the typical generator network and a syntactic image conditioned on image classes is generated.
	
	All the networks are implemented in chainer framework \cite{chainer_learningsys2015}. The standard  Adam optimizer with momentum is used for discriminator optimization, and the classifier network's convolutional layers weights are initialized using VGG 16-layer net pre-trained on ILSVRC dataset. For training the generator, we use Adam optimizer with isotropic Gaussian weights. Due to memory limitations, we use a batch of size 2; however, since the loss is computed for every pixel of training images and the final loss is averaged over those values, the batch-size is not that small. 
% 	For the first experimets we test our method on validation set of VOC 2012.
    We do not use any data augmentation or post-processing (e.g. CRF) in these experiments. 
	
%	\subsection{Inference}
	During testing, we only use discriminator network as our semantic segmentation labeling network. Given a test image, the softmax layer of the discriminator 
	outputs a set of probabilities of each pixel belonging to semantic classes, and accordingly, the label with the highest probability is assigned to the pixel. %One can use a post processing algorithms, for instance dense CRF, to improve further the results.
	%--------------------------------
	\section {Experimental Results}
	
	%\subsection{Implementation details}
	We evaluate our method on PASCAL VOC 2012 \cite{pascal-voc-2012}, SiftFlow \cite{liu2011sift},\cite{xiao2010sun}, StanfordBG \cite{gould2009decomposing}  and CamVid \cite{BrostowSFC:ECCV08} datasets.
	In the first experiment for Pascal dataset, we use all training data (1400 images) for which the pixel-level labels are provided as well as about 10k additional images with image-level class labels, i.e., for each image its semantic classes are known, but not the pixel-level annotations. These images are used in the weakly supervised setting. In the second experiment on Pascal dataset, for semi-supervised training, we use about $30\%$ (about 20 samples per class) of pixel-wise annotated data and the rest of images were without pixel-wise annotations.
	As metrics, we employ \textit{pixel accuracy}, which is per-pixel classification accuracy, \textit{mean accuracy}, i.e, average of pixels classification accuracies on number of classes  and \textit{mean IU}, average of region intersection over union (IU).
	
	% TODO: Describe briefly the employed metrics
	
	\begin{table}
		\centering
		\caption{The results on val set of VOC 2012 using all fully labeled and unlabeled data in train set. }
		\label{tabvoc1}
		\begin{tabular}{@{}llll@{}}
			\hline
			method      & pixel acc & mean acc & mean IU \\ 
			\hline

            % Full - baseline 1 decov & 89.6      & 65.0     & 56.3\\
% 			Weak Sup 1 decov & 91.6     & 77.5    & 66.5/z  \\ 
% 			\hline
			Full - our baseline & 89.9      & 69.2     & 59.5  \\
			Semi Supervised & 90.5      & 80.7     & 64.1  \\
			Weak Supervised & 91.3     & 80.0     & 65.8  \\ 
			
			\hline
			% 	ToDO	they had some overlap in train and val?
			FCN \cite{long2015fully} & 90.3      & 75.9     & 62.7  \\
			EM-Fixed \cite{papandreou2015weakly} & - & - & 64.6
% 			they did not provide that in the paper
	\vspace{-2mm}
		\end{tabular}
	\end{table}
	\begin{table}
		\centering
		\caption{The results on VOC 2012 validation set using $30\%$ of fully labeled data and all unlabeled data in training set. }
		\label{tabvoc2}
		\begin{tabular}{@{}llll@{}}
			\hline
			method      & pixel acc & mean acc & mean IU \\ 
			\hline
			Fully supervised & 83.15      & 53.1     & 38.9  \\
			Semi supervised & 83.6     & 60.0     &42.2  \\
			Weak Supervised & 84.6      & 58.6     & 44.6  
		\end{tabular}
			\vspace{-5mm}
	\end{table}

	% In Figure \ref {fig:errplot} we show the behavior of different losses.\\
	\begin{figure*}[t]
		\begin{center}
			\includegraphics[width=.87\textwidth]{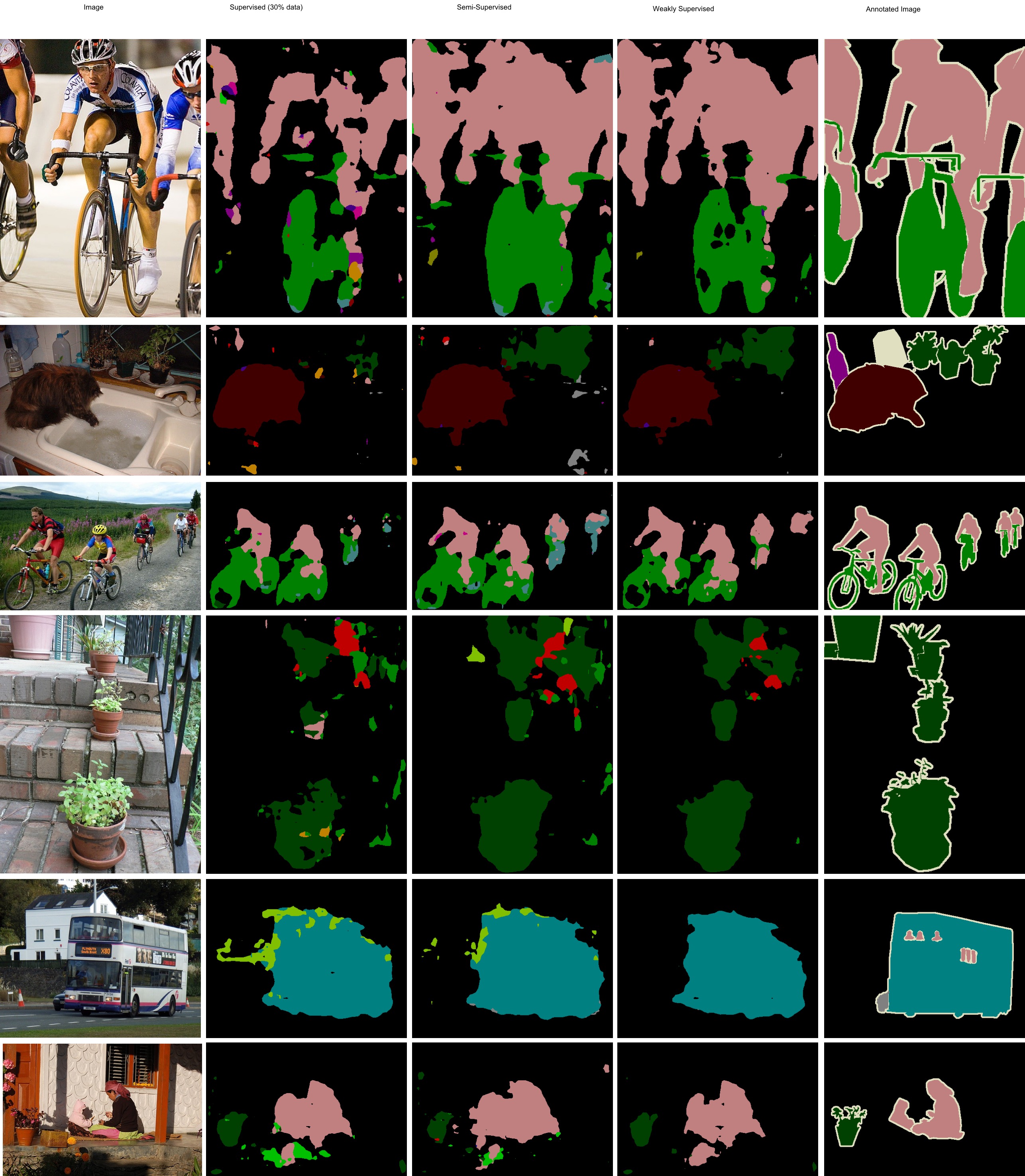}%\includegraphics[width=0.8\linewidth]{egfigure.eps}
		\end{center}
		\caption{Qualitative segmentation results on VOC 2012 validation set. The first to fifth  columns, respectively, show: the original images, the results of supervised learning using only 30\% of labeled data, the results of the proposed semi-supervised learning using 30\% labeled and about 400 unlabelled images, the results obtained using proposed weakly supervised learning with 30\% of labeled data and additional 10k images with image level class labels, and the Ground Truth. Both semi-supervised and weakly-supervised learning methods outperform the fully-supervised method. Weakly-supervised approach is more successful in suppressing false positives (background pixels misclassified as part of one of the $K$ available classes). }
% 			\vspace{-4mm}
		\label{fig:vocqul}
	\end{figure*}
% 	\vspace{-10mm}
	\begin{figure*}[t]
		\begin{center}
			\includegraphics[width=0.8\linewidth]{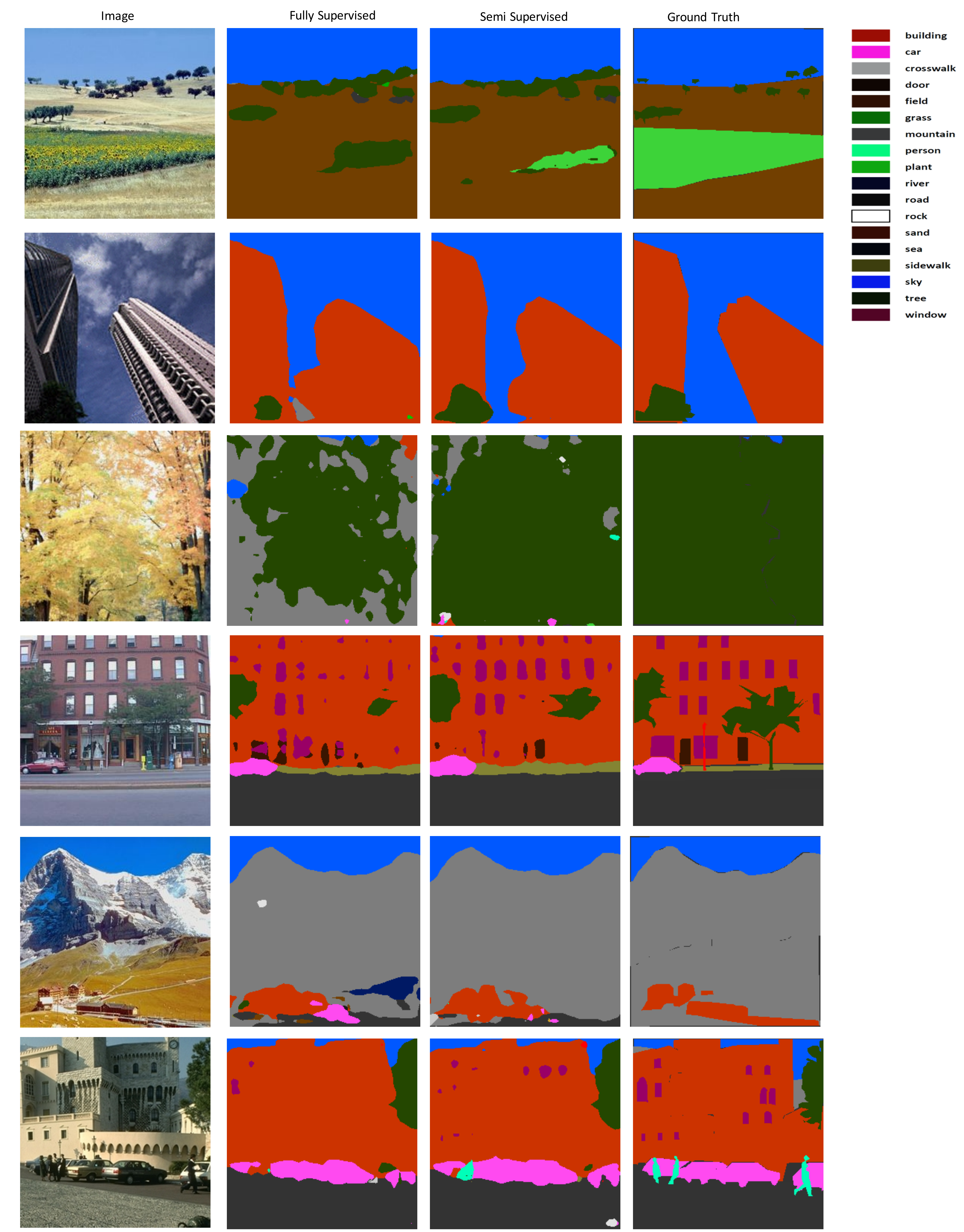}
		\end{center}
		\caption{Qualitative results on SiftFlow dataset, using unlabeled data results in more accurate semantic segmentation, unlikely classes in the image are removed in semi-supervised approach.}
% 			\vspace{-2mm}
		\label{figsiftres}
	\end{figure*}
		% TODO: channge this
		\begin{figure}[t]
		\begin{center}
			\includegraphics[width=.4\textwidth]{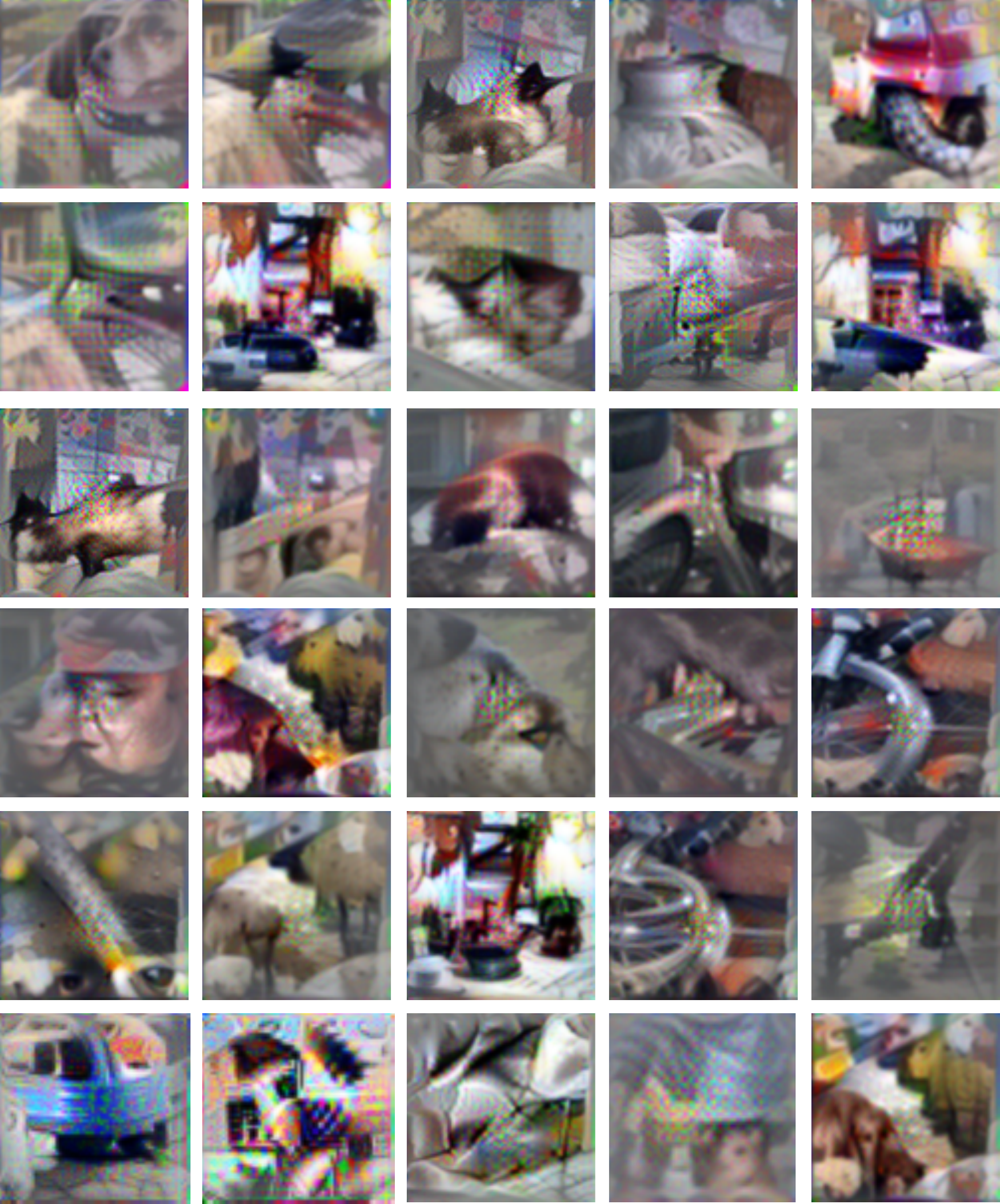}
		\end{center}
		\caption{Images generated by the generator of our conditional GAN on the Pascal dataset. Interestingly, patterns about dogs, cars, plants and cats have been automatically discovered. This highlights the effectiveness of our approach, indeed, the generator identifies automatically visual clusters that are then employed by the discriminator as pixel-level annotated data.}
		\label{fig:genWeakVoc}
	\end{figure}
		\begin{figure}[t]
		\begin{center}
			\includegraphics[width=.4\textwidth]{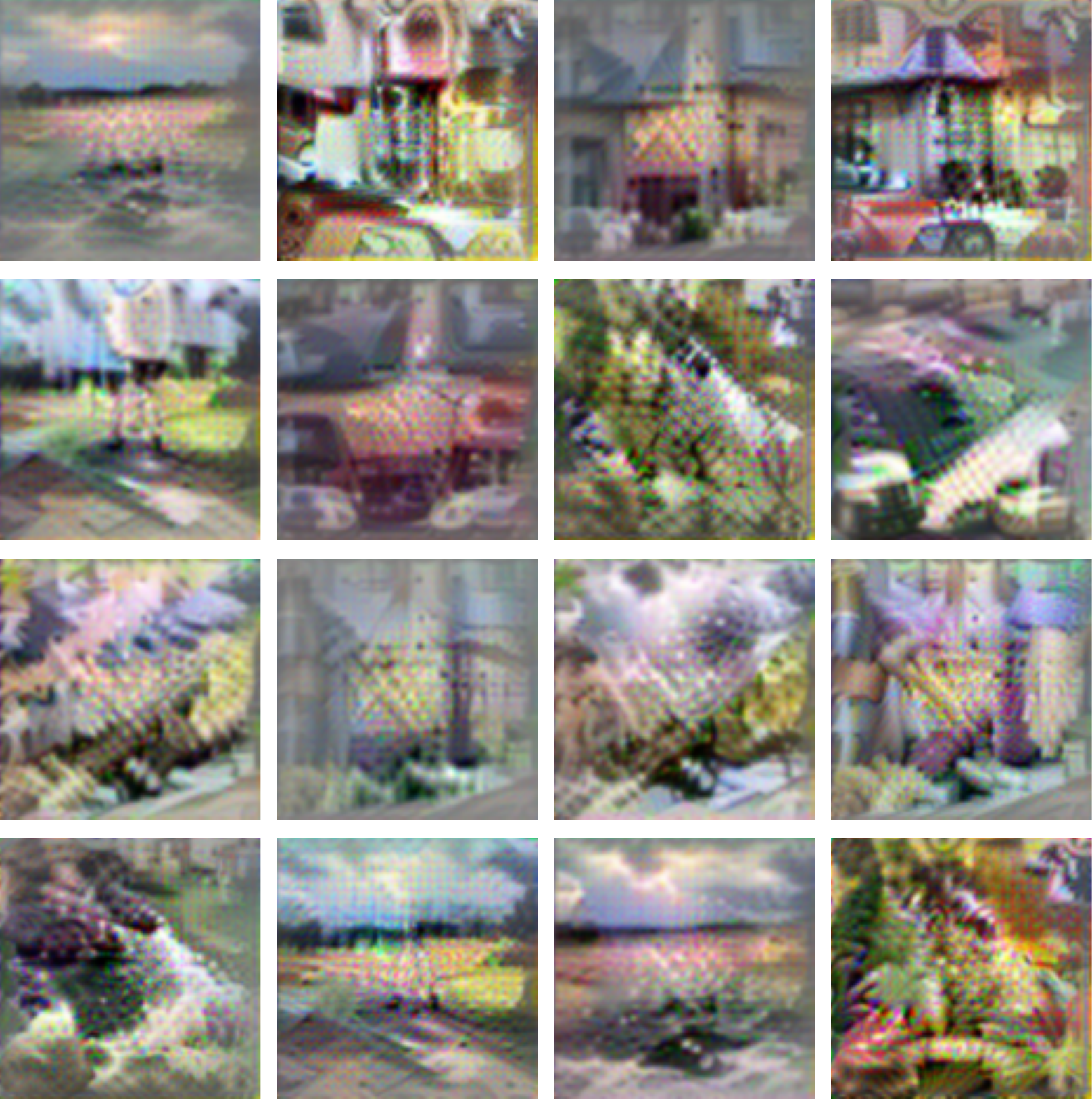}
		\end{center}
		\caption{Images generated by the generator during our GAN training on the SiftFlow dataset. Patterns about forests, beaches and slies can be observed.}
			\vspace{-4mm}
		\label{fig:genSiftflow}
	\end{figure}
	
		\begin{figure}[t]
		\begin{center}
			\includegraphics[width=.4\textwidth]{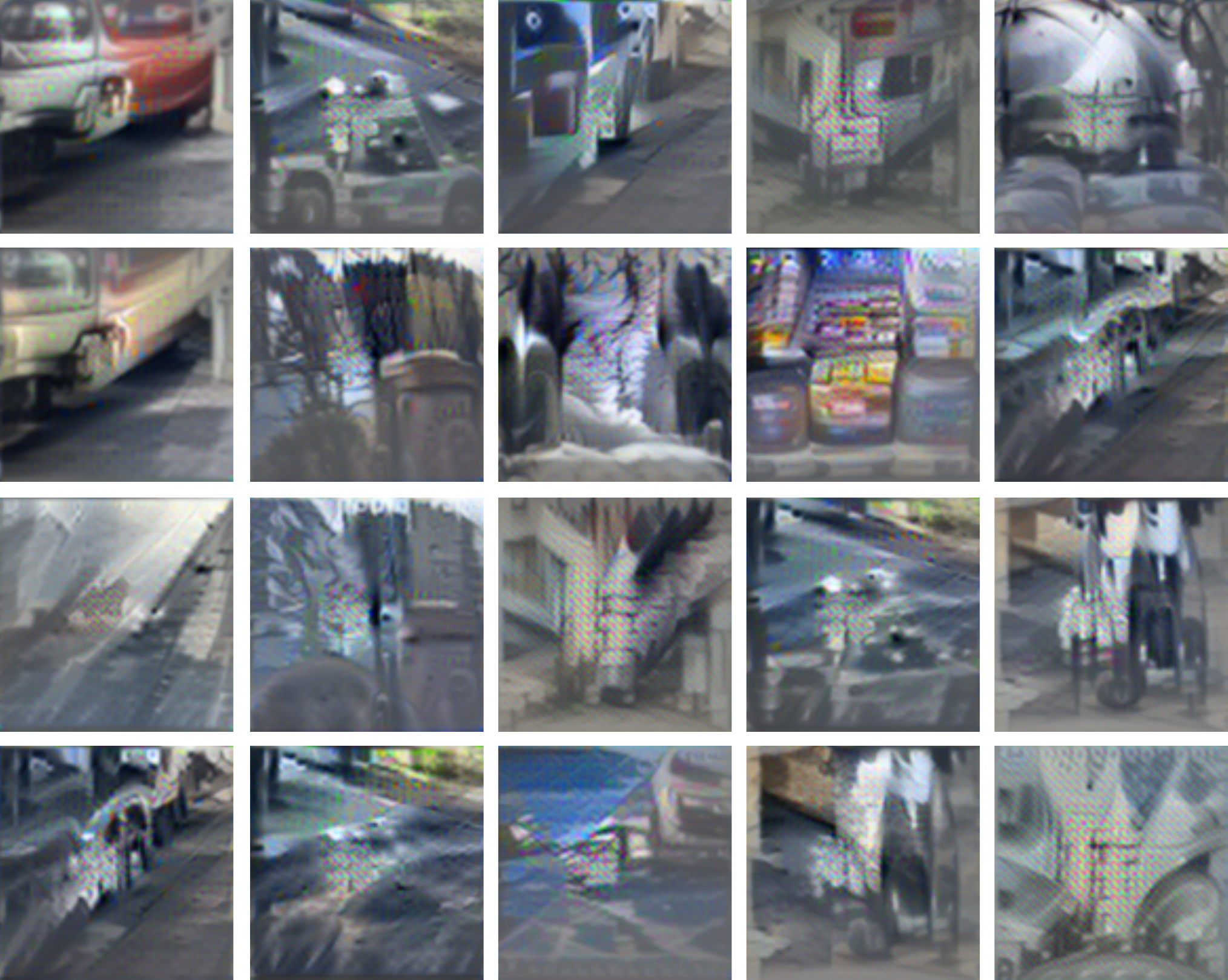}

		\end{center}
		\caption{Images generated by the Generator for the CamVid dataset. Patterns about mountains, cars and building can be observed. }
		
		\label{fig:genres}
	\end{figure}
% 		\vspace{-4mm}
	Quantitative results of our method on VOC 2012 validations set are shown in Tables \ref{tabvoc1} and \ref{tabvoc2}, and the qualitative results on some sample images are depicted in Figure \ref{fig:vocqul}. As shown in Table \ref{tabvoc2}, the semi-supervised method improves notably mean accuracy about $5\%$ to 7$\%$. The pixel accuracy is not significantly improved due to some false positives, which correspond to background pixels promoted by unlabeled data belonging to one of the classes in the training set. 
    False positives are reduced in the weakly supervised framework, due to the fact that the unsupervised loss encourages only labels occurring in the image and assigns them high confidences. This effect can be observed in qualitative results in Fig. \ref{fig:vocqul}. Thus, even though the semi-supervised method labels most of objects properly, it sometime assigns semantic classes to background pixels, while the weakly supervised method is able to reduce false positive detections. Furthermore, as shown in the same Tab. \ref{tabvoc1}, our weakly approach also outperforms state of the art semi-supervised semantic segmentation methods, such as \cite{papandreou2015weakly}, adopting a similar strategy to our weakly-supervised one. 
		
	Table \ref{tabsift} shows the results achieved by our approaches over the SiftFlow dataset \cite{liu2011sift}. Since in this dataset, background pixels are also labeled, the pixel accuracy is improved compared to the results obtained on PASCAL VOC 2012 dataset. 
	
		\begin{table}
		\caption{The results on SiftFlow using fully labeled data and 2000 unlabeled images from SUN2012 }
		\label{tabsift}
		\begin{tabular}{@{}llll@{}}
			\hline
			method      & pixel acc & mean acc & mean IU \\ 
			\hline
			Fully supervised & 83.4     & 46.7    & 34.4  \\
			Semi supervised & 86.3     & 50.8     & 35.1  
		\end{tabular}
% 			\vspace{-4mm}
	\end{table}
	Since images with class level labels are not available in the SiftFlow dataset, we only test semi-supervised learning. Fig. \ref{figsiftres} shows qualitative results on the SiftFlow dataset. In this case, unlabeled data allows us to refine the classification that initially are labeled with incorrect classes. For instance, in the fifth row the pixels which are mistakenly labeled as car or river are corrected in the semi-supervised results. Moreover, some small objects, such as the person or windows in the last row of Figure \ref{figsiftres}, which are not detected before, can be labeled correctly by employing additional data. 
	%  Samples from data generated by the generator network during the training Pascal dataset are shown in figure \ref{figgenres}. Unlike generated images from SiftFlow dataset, these images are not indicating specific objects when we first look at them; however, local regions can be interpreted as parts of objects for example wheels (bikes) or bird heads. One explanation can be that these images are generated from smaller labeled data, also variances of intra-class in Pascal are higher than Siftflow dataset.
	
	% \begin{figure*}[t]
	% \begin{center}
	%  \includegraphics[width=1\textwidth]{./gen_voc.pdf}\\
	%   %\includegraphics[width=0.8\linewidth]{egfigure.eps}
	% \end{center}
	%   \caption{  Generated images via generator in semi-supervised learning for PASCAL dataset  }
	% \label{figgenres}
	% \end{figure*}
	
	We repeated the semi-supervised experiments with different training set sizes e.g. 20\% and 50\% of labeled data, and the results are presented in Table \ref{tabpercent}. This results suggest that the extra data, in company with the way of the loss formulated, act as regularizer. Also, using more labeled data increases the overall performances, and the gap between the two settings is reduced.
	
		% TODO: move this to supplementary 
	\begin{table}
		\centering
		\caption{The results using different percentages of fully labeled data and all unlabeled data in train set. }
		\label{tabpercent}
		\begin{tabular}{@{}llll@{}}
			\hline
			method      & pixel acc & mean acc & mean IU \\ 
			\hline
			VOC $20\%$ Full & 73.15      & 23.2     & 16.0  \\
			VOC $20\%$ Semi  & 79.6     & 27.1     &19.8  \\
			\hline
			
			VOC $50\%$ Full & 88.5     & 63.6     & 51.6  \\
			VOC $50\%$ Semi  & 88.4     & 66.6     & 54.0  \\
			\hline
			
			SiftFlow $50\%$ Full & 79.0     & 28.3     & 21.0  \\
			SiftFlow $50\%$ Semi & 81.0    & 33.0     & 23.2  
			
		\end{tabular}
			\vspace{-3mm}
	\end{table}
	
	For the third experiment, we evaluated our method on StanfordBG \cite{gould2009decomposing} data set. This is a small data set including 720 labeled images, therefore we use Pasacal images as unlabeled data, since these images collected from pascal or similar datasets. Table \ref{tabstanford} shows our performance over the test images of StanfordBG data set compared to \cite{luc2016semantic}. It can be noted that our approach, again, outperforms significantly state of the art methods, e.g., \cite{luc2016semantic}, besides improving our fully-supervised method used as baseline.
	
	\begin{table}
		\caption{The results on StanfordBG using fully labeled data and 10k unlabeled images from PASCAL dataset}
		\label{tabstanford}
		\begin{tabular}{@{}llll@{}}
			\hline
			method      & pixel acc & mean acc & mean IU \\ 
			\hline
			Sem Seg Standard \cite{luc2016semantic}  & 73.3    & 66.5   & 51.3 \\
			Sem Seg Adv \cite{luc2016semantic} & 75.2     & 68.7   & 54.3 \\
			\hline
			Fully supervised & 77.5     & 65.1    & 53.1  \\
			Semi supervised & 82.3     & 77.6     & 63.3  
		\end{tabular}
			\vspace{-4mm}
	\end{table}
% 	\vspace{-2.5mm}

	Finally, we applied our proposed method to CamVid \cite{BrostowSFC:ECCV08} dataset. This dataset consists of 10 minutes videos (about 11k frames), for 700 images of which per-pixel annotations are provided. We use the training set of fully-labeled (11 semantic classes) data and all frames as unlabeled data, and we perform the evaluation on the test set.
% 	In this case, we use its 701 per-pixel annotated images of 11 semantic classes plus around 10k video frames available in the dataset 
We compare our results to SegNet \cite{badrinarayanan2015segnet} method in addition to our baseline (i.e., the fully-supervised method). The results are reported in Tab. \ref{tabCamvid} and show that our semi-supervised method notably improves per-class accuracy, which indicates that more present classes in the images are identified correctly. 
	\begin{table}
		\caption{The results on CamVid using fully labeled training data and 11k unlabeled frames from its videos. }
		\label{tabCamvid}
		\begin{tabular}{@{}llll@{}}
			\hline
			method      & pixel acc & mean acc & mean IU \\ 
			\hline
			Segnet-Basic \cite{badrinarayanan2015segnet}  & 82.2   & 62.3   & 46.3 \\
			SegNet (Pretrained) \cite{badrinarayanan2015segnet} & 88.6    & 65.9   & 50.2 \\
			\hline
			Ours Fully supervised & 88.4     & 66.7    & 57.0  \\
			Ours Semi supervised & 87.0     & 72.4     & 58.2  
		\end{tabular}
			\vspace{-5mm}
	\end{table}
	Samples of images generated by our GAN during training over the employed datasets are shown in Figures \ref{fig:genSiftflow}, \ref{fig:genres} and \ref{fig:genWeakVoc}. These images clearly indicate that our network is able to learn hidden structures (specific of each dataset) that are then used to enhance the performance of our GAN discriminator as they can be seen as additional pixel-level annotated data.  
	%We can see some classes such as sky, roads, trees and cars properly present in the images. %TODO:WHY??
	Moreover, interestingly, our GAN framework is also able to learn spatial object distributions, for example, roads are at the bottom of images, sky and mountains are at the top, etc.
	
\begin{figure*}[t]
\begin{center}
 \includegraphics[width=1\textwidth]{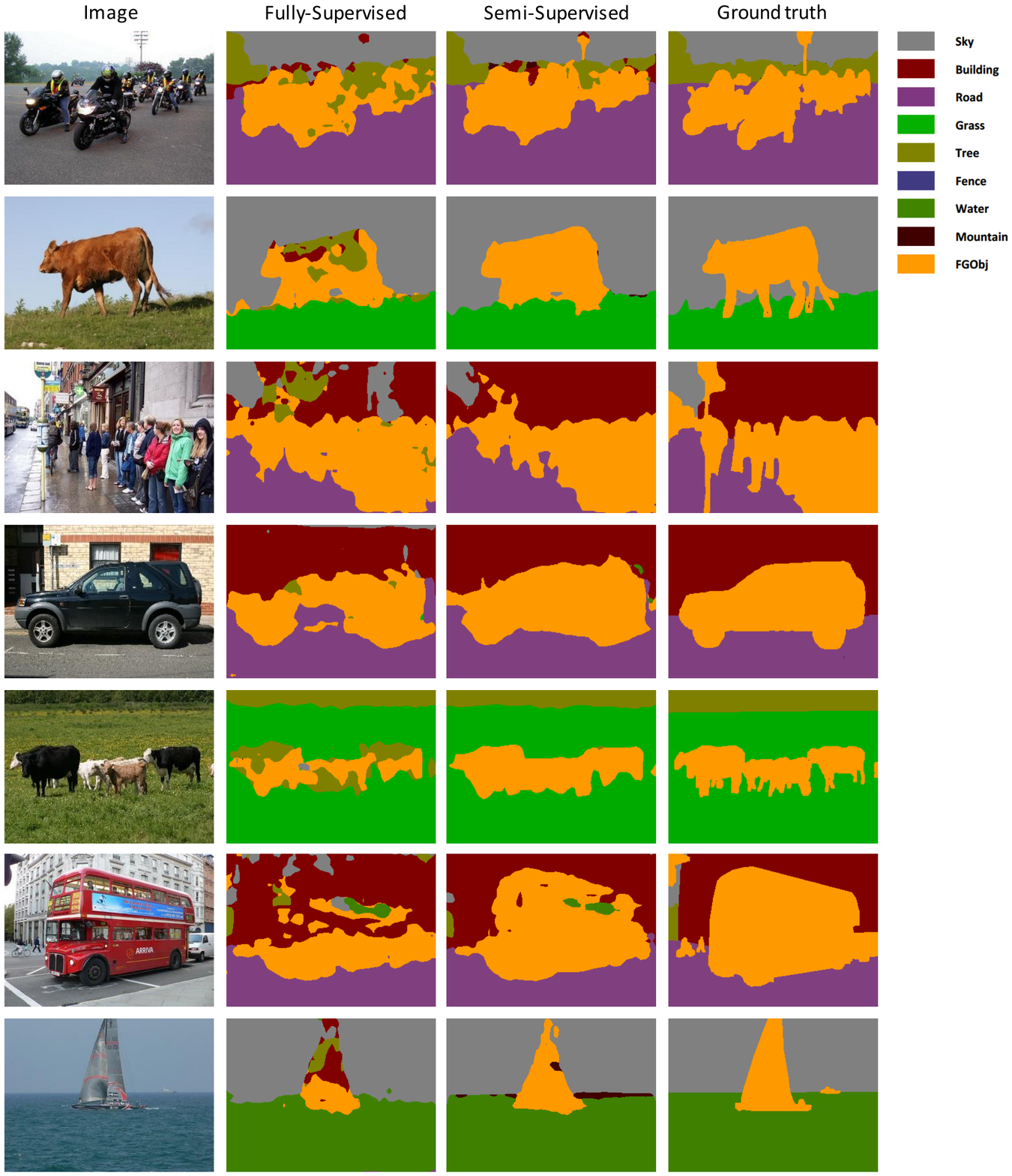}
\end{center}
  \caption{ Qualitative results on StanfordBG dataset, using unlabeled data results in more accurate semantic segmentation, unlikely classes in the image are removed in semi-supervised approach.  }
\label{figStanford1}
\end{figure*}

\begin{figure*}[t]
\begin{center}
 \includegraphics[width=1\textwidth]{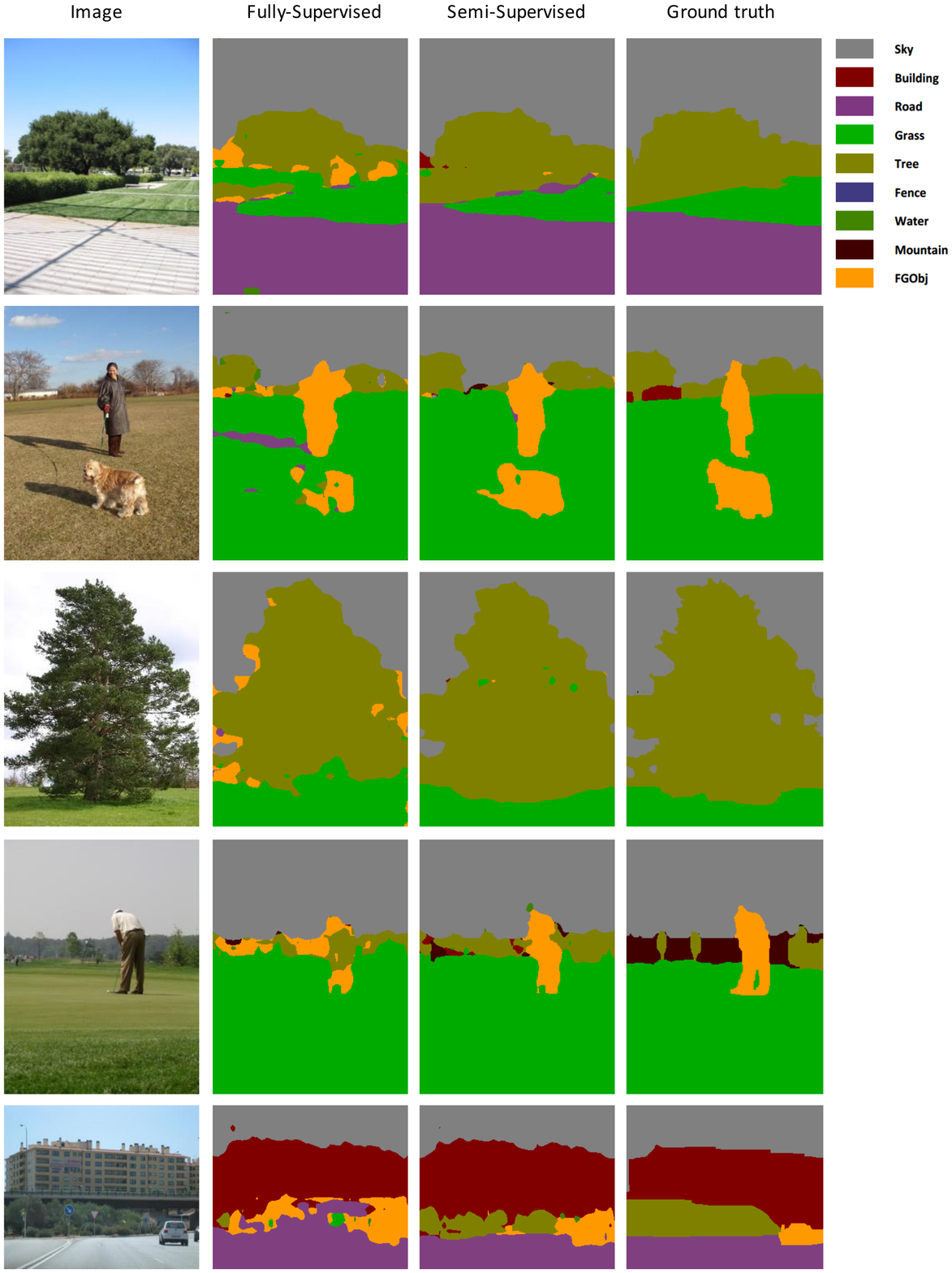}
\end{center}
  \caption{ More qualitative results on StanfordBG dataset. }
\label{figStanford2}
\end{figure*}

	In Figures \ref{figStanford1} and  \ref{figStanford2} examples from qualitative results for StanfordBG dataset are depicted; by using unlabeled data via our proposed approach some pixels which fully-supervised method labeled incorrectly, can be refined. For instance, in the second row parts from Cow which are mistakenly labeled as building or tree are corrected in the semi-supervised result. \\

\begin{figure*}[t]
\begin{center}
 \includegraphics[width=1\textwidth]{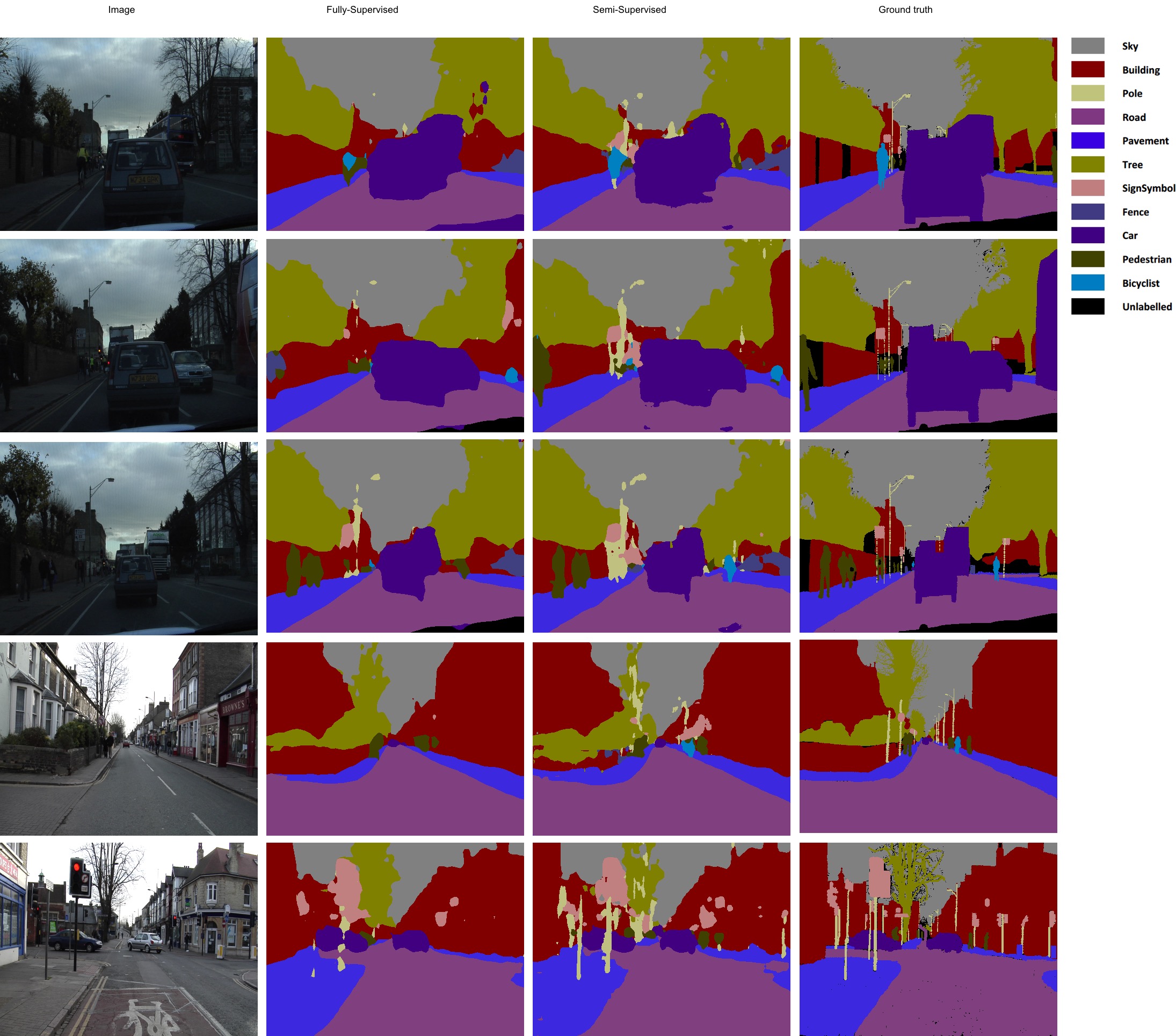}
\end{center}
  \caption{ Samples of qualitative results from CamVid dataset. More classes are captured in semi-supervised learning approach. }
\label{figCamvid}
\end{figure*}

Samples of qualitative results from CamVid dataset are shown in Figure \ref{figCamvid}.
As before, some pixels are refined using unlabeled data.
Moreover, some small objects, for example the pole, pedestrian or bicyclist in the figure \ref{figCamvid}, which are not detected can be labeled correctly by employing additional data.

	Summarizing, the  results achieved over different experiments indicate that the extra data provided through adversarial loss boosts the performance (outperforming both fully-supervised and state-of-the-art semi-supervised methods) of semantic segmentation, especially in terms of mean accuracy measure. The competitiveness of the discriminator and the generator yields not only  in generating images, but, most importantly,  to learn more meaningful features for pixel classification.
	
	\section {Conclusion}
	In this work, we have developed a novel semi-supervised semantic segmentation approach employing Generative Adversarial Networks. We have also investigated GANs conditioned by class-level labels, which are easier to obtain, to train our fully-convolutional network in a weakly supervised manner. We have demonstrated that this approach outperforms fully-supervised methods trained with a limited amount of labeled data as well as state of the art semi-supervised methods over several benchmarking datasets. 
	Beside, our model generates plausible synthetic images, which show some meaningful image features such as edges and class labels, that supports the discriminator in the pixel-classification step. The discriminator can be replaced by any better classifier suitable for semantic segmentation for further improvements. %We tested our model on VOC 2012, SiftFlow, StanfordBG and CamVid datasets and reported the improved results over the standard classifier results.
	
	{\small
		\bibliographystyle{ieee}
		\bibliography{semiSeg17}
	}
	
\end{document}